# *Replication Study: Enhancing Hydrological Modeling with Physics-Guided Machine Learning*

Mostafa Esmaeilzadeh[1], Melika Amirzadeh[2]

[1]Department of Civil Engineering, Shahrood University of Technology, Iran

[2]Department of Civil Engineering, Islamic Azad University of Mashhad, Iran

## Abstract

**Current hydrological modeling methods combine data-driven Machine Learning (ML) algorithms and traditional physics-based models to address their respective limitations—incorrect parameter estimates from rigid physics-based models and the neglect of physical process constraints by ML algorithms. Despite the accuracy of ML in outcome prediction, the integration of scientific knowledge is crucial for reliable predictions. This study introduces a Physics Informed Machine Learning (PIML) model, which merges the process understanding of conceptual hydrological models with the predictive efficiency of ML algorithms. Applied to the Anandapur sub-catchment, the PIML model demonstrates superior performance in forecasting monthly streamflow and actual evapotranspiration over both standalone conceptual models and ML algorithms, ensuring physical consistency of the outputs. This study replicates the methodologies of Bhasme, P., Vagadiya, J., & Bhatia, U. (2022) from their pivotal work on Physics Informed Machine Learning for hydrological processes, utilizing their shared code and datasets to further explore the predictive capabilities in hydrological modeling.**

## Introduction

Streamflow describes the volume of available freshwater and the flood discharge. Therefore, accurate streamflow prediction is important for water resource management (Cho & Kim, 2022, Moghaddasi et al., 2022, Sohrabi et al., 2022). The challenges of predicting streamflow and preventing flooding are still important for societal security (Niu & Feng, 2021). Ground stations collect the most reliable and accurate environmental data, but they often lack the detail and coverage needed for scientific studies and decision-making (Mohammadpouri et al., 2023, Karamouz et al., 2022b). This issue persists despite advancements in sensor networks and remote sensing technology (Willard et al., 2023). To address this issue, many researchers have turned to machine learning algorithms (Mohammadpouri et al., 2023, Fereshtehpour et al., 2024, Karamouz et al., 2022a). These methods are employed to predict how catchments will respond to meteorological forcings. However, due to their limited understanding, the effectiveness of machine learning (ML) techniques in practical applications remains challenging, characterized by physical inconsistencies and persistent issues with equifinality (Feng et al., 2020).

Physics-based models, utilizing equations to represent domain knowledge, remain scholars' and practitioners' preferred tool for addressing interpretability in scientific studies. Due to the rising recognition that neither only ML algorithms nor physics-based models are adequate to solve domain-specific difficulties, physics-guided machine-learning approaches are receiving substantial attention in the scientific and engineering fields (Vieux et al., 2004).

The literature has been devoted to fusing physics with machine learning approaches for a variety of applications, such as reconstructing discrepancies in the Reynolds model. Wang et al., 2017 suggested a data driven, PIML approach to various flow conditions. Muralidhar et al., 2019 revealed how to forecast the drag force which impacts every particle in fluid flow using a physics-guided model design. Zhang et al., 2020 developed a physics-guided Convolutional Neural Network to predict seismic response in buildings, highlighting the necessity for scalable data science methods integrating physical principles for earth and atmospheric research.

Karpatne et al. (2017) and Jia et al. (2019) created ML models that are physics-guided, which direct neural network architectures. Karpatne et al. (2017) specifically showed how to improve performance over models that are solely knowledge- or data-based by combining the common models for neural networks that incorporate energy conservation regulations. Jia et al. (2020) present a global Physics Guided Recurrent Graph Networks (PGRGrN) to predict flow and temperature in observed and unobserved river networks. Another research simulated the physics-based model, built a database, and analyzed the database using various data-driven methodologies Reference (Liang et al., 2019). Lu et al. (2021) proposed a physics-informed LSTM model to improve prediction when data is not widely available. The PRMS model was used to simulate streamflow and meteorological features. Khandelwal et al. (2020) suggested using an LSTM-based architecture to forecast streamflow. To model the streamflow, a second LSTM was fed with the first layer's outputs and the original SWAT inputs. Xu et al. (2014) combines instance-based weighting and support vector regression to enhance prediction accuracy in physically based regional groundwater flow models in the Republican River basin, USA.

This study develops a framework to integrate a conceptual hydrological model with cutting-edge ML models to leverage the predictive power of ML algorithms and the process comprehension of physics-based models in a complementary manner. The input variables (PET, P, GW, and SM), intermediate variables (actual ET), and goal variables (Q at a certain gauge site) are all identified using the model. To determine the connections between input and output variables (both intermediate and target), machine learning (ML) algorithms are substituted for empirical equations in various steps of the model. PIML models generalize basin characteristics, unlike empirical equations in conceptual hydrological models, which are basin-specific.

The manuscript is organized as follows: First, a short description of the abcd model, and the machine learning algorithms utilized in this work is presented. Next, the suggested detailed PIML model, as well as the various assessment criteria employed is covered. Additionally, it outlines the case study and datasets. The next section discusses the model settings for all models. After that, the ML and PIML models using the metrics are compared. Lastly, the conclusion of the work is presented.

**Methods:**

**abcd model**

The abcd hydraulic framework is a hydrological model that consists of two parts: the earth's aquifer and the underground water layer. It is based on the principle of water balance equation ($SM_t + ET_t + DR_t + GR_t = SM_{t-1} + P_t$). Here, $P_t$ represents monthly precipitation, $ET_t$ represents the actual monthly vaporization, $DR_t$ is the direct surface overflow, $GR_t$ symbolizes groundwater recharge, and $SM_t$ and $SM_{t-1}$ indicate the soil moisture content of the present and previous months, respectively (Yue et al., 2023). The framework divides the total runoff into three components: surface runoff, interflow, and baseflow, and requires the calibration of nine parameters to simulate streamflow accurately. The abcd framework is efficient in simulating evapotranspiration flows and outflow at the basin level, capturing the crucial hydrological processes that contribute to water balance (Picourlat et al., 2022).

In this study, we have used the abcd hydrological model to determine how water flows (Q) in response to precipitation (P) and potential evaporation and transpiration (PET). This model considers two storage areas: ground water (GW) and soil moisture (SM), which is a pivotal environmental factor affecting soil evaporation and transpiration (Hosseini et al., 2023).

Evapotranspiration (ET), surface runoff, and groundwater recharge (GR) all contribute to the loss of moisture from the soil, which is replenished by precipitation. Water is added to the groundwater compartment through recharge and removed through discharge (GD). The total streamflow is the result of surface runoff and groundwater discharge. Fig. 1 conceptually illustrates the abcd model.

The model requires a time series of precipitation, air temperature, and streamflow data. The approach outlined here is used to compute PET using air temperature data. The behavior of the model is determined by four factors: a regulates the amount of recharge and runoff that takes place when the soils are not sufficiently saturated; b controls the soil's saturation level; c specifies the proportion of surface runoff to groundwater recharge; and d regulates the groundwater outflow rate.

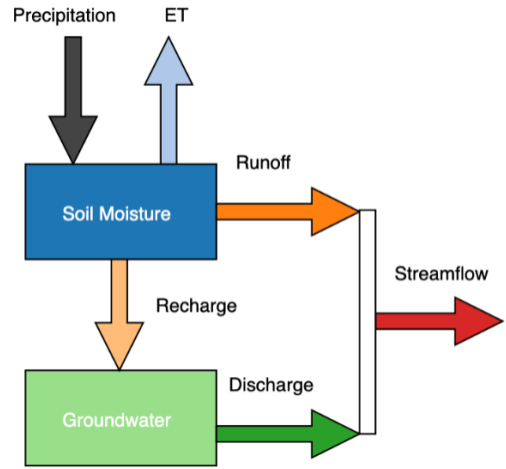

**Fig. 1.** abcd model

Fig. 2 depicts the interactive visualizations with values of 0.93, 5, 0.4, and 1.5 for a, b, c, and d, respectively. It shows how the model calculates the fluxes and storage terms for every time step. The mass balance in Equation 1 forms the basis for the diagrams.

$$\text{Storage}_t + \sum \text{Outflow}_t = \text{Storage}_{t-1} + \sum \text{Inflow}_t \quad (1)$$

According to this equation, the initial storage plus the total inflow must equal the sum of the remaining storage plus the total outflow.

For the soil moisture section, the mass balance equation and the other relevant equations are shown in Equations 2 to 8.

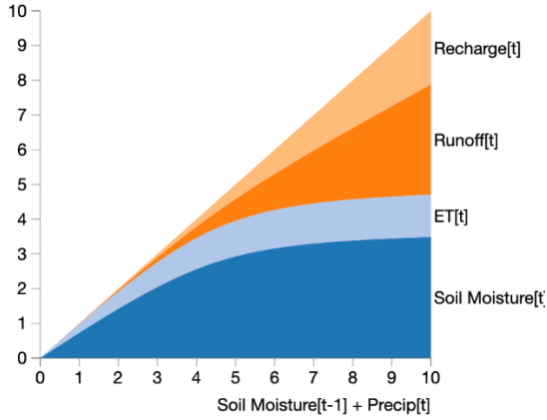

**Fig. 2.** Interactive visualizations of four parameters.

$$SM_t + ET_t + DR_t + GR_t = SM_{t-1} + P_t \quad (2)$$

$$W_t = SM_{t-1} + P_t \quad (3)$$

$$Y_t = SM_t + ET_t = \frac{W_t+b}{2a} - \sqrt{\left(\frac{W_t+b}{2a}\right)^2 - \frac{b \cdot W_t}{a}} \quad (4)$$

$$ET_t = Y_t \times (1 - e^{-PET_t/b}) \quad (5)$$

$$SM_t = Y_t \times e^{-PET_t/b} \quad (6)$$

$$DR_t = (1-c) \times (W_t - Y_t) \quad (7)$$

$$GR_t = c \times (W_t - Y_t) \quad (8)$$

For the groundwater section, the mass balance equation and the other relevant equations are shown in Equations 9 to 12.

$$GW_t + GD_t = GW_{t-1} + GR_t \quad (9)$$

$$GW_t = \frac{(GW_{t-1} + GR_t)}{1+d} \quad (10)$$

$$GD_t = d \times GW_t \quad (11)$$

$$Q_t = DR_t + GD_t \quad (12)$$

### Review of Machine Learning Algorithms

In recent years, machine learning techniques have seen growing utilization within hydrological studies, showcasing considerable promise in enhancing the precision of simulation and prediction procedures (Wang et al., 2023). These methodologies, adept at discerning patterns from intricate geospatial and hydrological datasets, have exhibited commendable performance (Lange and sippel, 2020; Reichstein et al., 2019). The foundation of the suggested PIML framework lies in machine learning algorithms. To enhance comprehension of subsequent sections, a review of the diverse methodologies employed in this investigation is provided. Fig. 3 depicts the machine learning workflow implemented in this study.

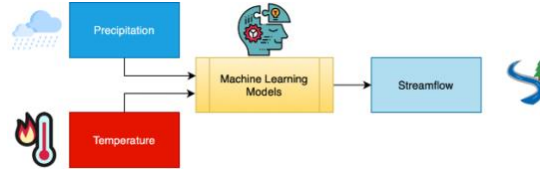

**Fig. 3.** Machine learning workflow

### LSTM

LSTM networks are widely used in hydrological modeling due to their ability to understand long-term connections in time series data. This is particularly useful when analyzing phenomena that involve temporal dynamics, such as streamflow patterns and the relationship between rainfall and runoff. LSTMs have been successfully applied to various hydrological scenarios, including predicting streamflow, establishing links between anomalies in water table depth and precipitation, and modeling runoff. Research studies conducted by Tan et al. (2023), Latifoğlu (2022), Ma et al. (2021), and Hashemi et al. (2022) have demonstrated that LSTMs are remarkably effective compared to conventional machine learning models. LSTMs have shown their capability to capture the intricate and nonlinear relationships that underlie hydrological processes.

### GPR

Gaussian Process Regression (GPR) is a regression method that doesn't rely on a specific mathematical model. Instead, it describes a probability density distribution over a range of potential functions that fit a particular set of data points. This distribution can be updated with new data points as they become available. At the core of GPR are the conditional probability distribution and covariance matrix between the observed and unobserved data points (Ramezani et al., 2023). The probability distribution can be used to derive the mean and confidence intervals, which are then used to predict and measure the uncertainty inherent in the prediction. GPR has been used by researchers to forecast streamflow with success in the past (Sun et al., 2014).

### SVR

Support Vector Regression (SVR) is commonly utilized in various research fields, including classification of remotely sensed images (Mohamadzadeh and Ahmadisharaf, 2024), and is also popular in hydrology for regression tasks, particularly in support vector machine regression (SVMR) analysis (Liu et al., 2022). SVRs are effective in learning non-linear correlations between input and output by using the kernel approach. This approach converts the inputs into feature spaces with high dimensions that can be solved linearly (Malik et al., 2022).

**LASSO and Ridge**

Regularization methods like Ridge regression and the least absolute shrinkage and selection operator (LASSO) are used in high-dimensional problems to improve model performance by avoiding overfitting (Djeundje et al., 2021). Ridge regression adds a small degree of bias to regression predictions by using an L2 regularization approach that aims to achieve more accurate outcomes (Satpathi et al., 2023). On the other hand, LASSO estimates regression coefficients by maximizing the penalized log-likelihood with an L1-norm regularization function that simplifies complexity through regularization. The outputs of LASSO and Ridge regressions have been found to be competitive with cutting-edge ML techniques (Karpatne et al., 2017).

**Physics Informed Machine Learning model**

The suggested Physics Informed Machine Learning (PIML) model is a way to combine a physics-based conceptual model with machine learning techniques to improve performance. The model uses the physics-based conceptual model (in this case, the abcd model) and replaces hard mathematical correlations between input and output variables with machine learning algorithms. This allows for the identification of complex relationships between input and output variables, while still maintaining the interpretability and physics-informed selection of covariates provided by the conceptual model.

In this study as it can be shown in Fig. 4, predictors such as PETt, Pt, and SMt-1 are used to find the predictand, ETt. The estimates of ETt, along with SMt, SMt-1, GWt, GWt-1, and Pt are then combined to create a new covariate matrix. This matrix is then sent into the next layer of the machine learning algorithm to get the target variable, Qt.

The general functional connection between ETt and Qt can be expressed as follows.

$$ET_t = f(SM_{t-1}, P_t, PET_t) \quad (13)$$

$$Q_t = g(SM_{t-1}, SM_t, GW_t, GW_{t-1}, P_t, E_t \quad (14)$$

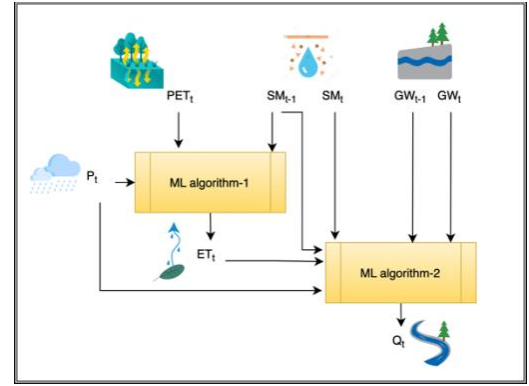

**Fig. 4.** Workflow of PIML model for streamflow prediction

**Evaluation metrics**

Nash-Sutcliffe Efficiency (NSE), Percent Bias (PBIAS), and Root Mean Square Error (RMSE) are utilized to evaluate the performance of the models. These metrics, which are employed in some hydrological applications (Najafi et al., 2016) (Wagena et al., 2020), evaluate the model's effectiveness, biases in its predictions, and estimate mistakes in its outputs, respectively.

$$NSE = \sqrt{\frac{\sum_{n=1}^{i}(S_i - O_i)^2}{\sum_{n=1}^{i}(S_i - \bar{O}_i)^2}} \quad (15)$$

$$PBIAS = \frac{\sum_{i=1}^{n}(S_i - O_i)^2}{\sum_{i=1}^{n} O_i^2} \quad (16)$$

$$RMSE = \sqrt{\frac{\sum_{i=1}^{n}(S_i - O_i)^2}{n}} \quad (17)$$

$O_i$, $S_i$, and $\bar{O}$ are the 'observed', 'simulated', and mean of observed values, respectively.

**Study area and datasets.**

To show the implementation of the methodology, as it can be shown in Fig. 5, the Anandapur sub-catchment is selected which is in the Baitarni basin with an 8667.95 km² area. The India Meteorological Department (IMD) provided the temperature and precipitation data at a spatial resolution of 1° and 0.25°, respectively, for the years 1979 through 2014.

The India Water Resources Information System is where the observed streamflow data is found. The Global Land Data Assimilation System (GLDAS) Catchment Land Surface Model L4 daily datasets are used to calculate groundwater storage, evapotranspiration, and daily soil moisture.

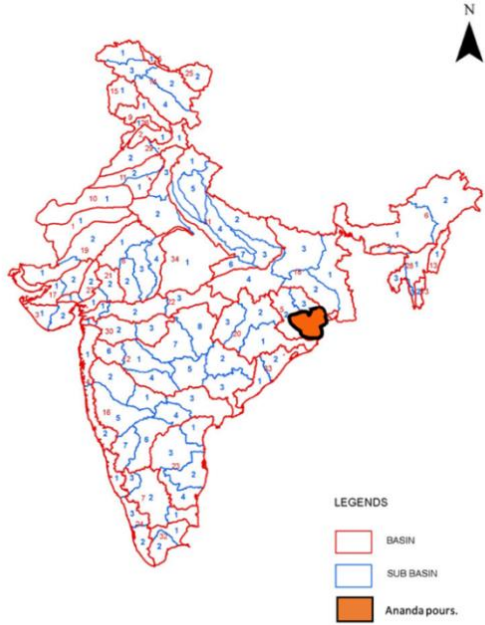

**Fig. 5.** Location of Anandapur sub-catchment in India.

To evaluate model performance, different training and test periods are considered to ensure the test dataset does not have instances from the training datasets. 1979-2008 are used for training and 2009-2014 for test. These data are the same for the abcd, ML, and PIML models, shown in Table 1.

**Table 1.** Datasets utilized in this work

| Data | Spatial Resolution | Source |
|---|---|---|
| Precipitation | 0.25˚ | IMD |
| Temperature | 1 | IMD |
| Soil moisture, Groundwater storage and actual evapotranspiration | 0.25˚ | GLDAS |
| Streamflow | Gauge | India-WRIS |

## Implementation and Results

### Machine Learning Methods

For all the models, we programmed in Python using the Keras, Tensorflow (version 2.10.1), and Sklearn libraries, and we implemented them all in Google Colab. we utilized the Sklearn package and associated functions to create the ML models, and we used the grid search tool to identify the ideal hyperparameters for either ML or PIML implementation.

To put the LSTM model into practice, we utilized the Keras and Tensorflow libraries. We compiled it using the MSE loss function with ADAM optimizer and utilized one LSTM, dropout, and dense layer. To cut down on the running time, we also employed an early stop. we also used the Minmax function to standardize the inputs before applying the LSTM model, and then applied it once more to invert them.

### Performance valuation of ML Algorithms

The performance of different ML algorithms for streamflow prediction is presented in this section. These algorithms are connected to the conceptual hydrological model. Precipitation and temperature are used as inputs of ML algorithms. Performance metrics of ML algorithms are presented in Table 2. In this case, five different ML models including LSTM, LASSO, Ridge, SVR, and GPR are applied that LSTM and GPR show satisfactory performance, while the rest of the models' performance is in the unsatisfactory range. For LSTM and GPR the RMSE values are less and the NSE values are higher than the other models. Fig. 6 shows the time series of observed and best-predicted streamflow models (LSTM and GPR) in the test period (2009-2014) with monthly timestep. The LSTM has a slightly better performance rather than the GPR model. And it should be noted that LSTM prediction for the peak's values is satisfactory.

Even though different ML models may consider the non-linear relationship between inputs and outputs, it is observed that it is challenging to understand why certain predictions are made. This problem is addressed in this paper by simulating intermediate processes using the suggested PIML model and embedding ML methods.

**Table 2.** Performance evaluation of ML models

| ML MODELS | | | |
|---|---|---|---|
| Variable | Q | | |
| Performance metric | RMSE | PBIAS | NSE |
| LSTM | 40.719 | -4.841 | 0.636 |
| LASSO | 43.219 | -15.048 | 0.586 |
| Ridge | 43.219 | -15.05 | 0.585 |
| SVR | 45.157 | 4.470 | 0.548 |
| GPR | 41.415 | -4.733 | 0.619 |

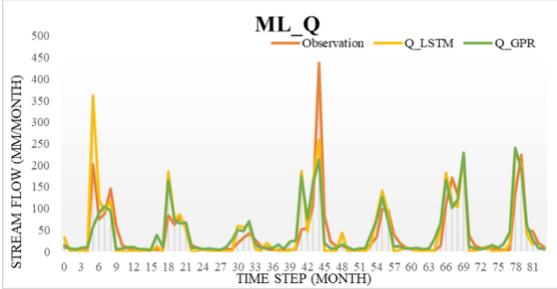

**Fig. 6.** Time series of observed and predicted streamflow of LSTM and GPR in the test period.

## Performance Evaluation of PIML model

The performance of the PIML model for streamflow (Q) and evapotranspiration (ET) during the test period is evaluated in this section (Fig. 7 and Fig. 8). Table 3 shows the results of the PIML model with five different ML models. The NSE values of streamflow demonstrate that the PIML model with LSTM performs satisfactorily while other models' performance is not very good. LSTM has the best performance in evapotranspiration prediction during the period of this study. It has the highest value of NSE (0.768) and the lowest RMSE (11.654) rather than the others. In addition, there is an improvement in NSE values obtained from streamflow in most of the models from ML models to PIML models.

**Table 3.** Performance assessment of PIML models during the test period

| PIML MODELS | | | | | | |
|---|---|---|---|---|---|---|
| Variable | ET | | | Q | | |
| Performance metric | RMSE | PBIAS | NSE | RMSE | PBIAS | NSE |
| LSTM | 11.654 | 1.889 | 0.768 | 36.778 | -23.694 | 0.703 |
| LASSO | 16.940 | -1.746 | 0.510 | 40.031 | -21.971 | 0.648 |
| Ridge | 16.940 | -1.749 | 0.510 | 40.217 | -21.830 | 0.645 |
| SVR | 14.489 | 1.024 | 0.642 | 59.632 | 53.967 | 0.219 |
| GPR | 14.725 | 1.817 | 0.630 | 43.819 | -13.150 | 0.578 |

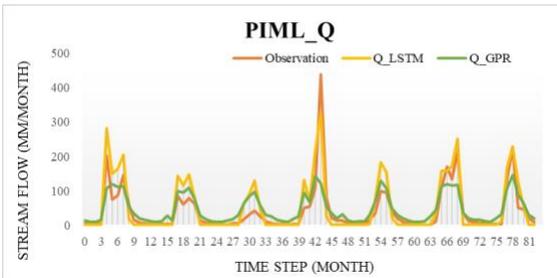

**Fig. 7.** Time series of observed and predicted streamflow of the PIML model considering the LSTM and GPR in the test period.

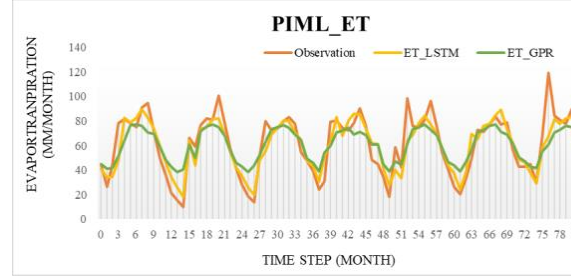

**Fig. 8.** Time series of observed and predicted evapotranspiration of the PIML model considering the LSTM and GPR in the test period.

## Discussion and Conclusion

Over the last few years, researchers have been taking advantage of deep learning methods to rainfall-runoff modeling which led to enhancing the performance of physical-based models. The main reason is the theoretical discovery of which training model can be helpful to figure out the rules and policies of water resources systems (Karpatne et al., 2017). Deep learning methods often used to model rainfall-runoff include Long Short-Term Memory (LSTM) networks, which are types of recurrent neural networks (RNN). The LSTM is capable of modeling rainfall-runoff cycles effectively because it captures long-term dependencies. Moreover, it can be combined with other traditional models to capture the nonlinearity and uncertainty of the physical process. The overall performance of rainfall-runoff models can be improved. Therefore, deep learning models, especially LSTM, have great potential for application in rainfall-runoff modeling (Xie et al., 2021).

A PIML model is proposed here for predicting both target and intermediate variables. The PIML model is based on a combination of the three key components of data-driven modeling: artificial neural networks, evolutionary algorithms, and Bayesian networks. Hydrological processes rely on two key variables, actual evapotranspiration, and streamflow, which the model directs to predict accurately. An example of how the model can be utilized in a single hydrological unit has been provided (Xiong et al., 2019).

During this introduced model (PIML), the physical interpretation of hydrological models has been integrated with machine learning algorithms for the first time. This combination provides a reasonable way to interpret the model outputs. Further, the PIML model quantifies uncertainties in both intermediate and target variables in a given case study. In this study, the outputs from the PIML model showed high prediction accuracy and were able to capture the underlying patterns in the data. This makes it an excellent tool for hydrological forecasting and prediction. As a result, the PIML model can be a useful tool for hydrologists and researchers to better understand and interpret hydrological systems.

It is necessary to validate future extensions to the PIML framework, including distributed and semi-distributed models, and daily and sub-daily time-steps. To achieve this purpose, we should investigate how upstream reservoirs work when it is a part of the model's predictive skills. Furthermore, the PIML's accuracy should be evaluated for various input datasets, including those with higher temporal resolution, such as hourly or sub-hourly data. PIML models can also be tested against other model types to compare their relative performance in predicting hydrological processes. Finally, further research into the best ways of combining the PIML model with existing hydrological models to enhance their predictive capabilities should be conducted.

Our findings build upon the foundational work of Bhasme, P., Vagadiya, J., & Bhatia, U. (2022) in "Enhancing predictive skills in a physically-consistent way: Physics Informed Machine Learning for hydrological processes," demonstrating the robust applicability of their methods and datasets in advancing the field of hydrological modeling. This study not only replicates their innovative approach but also extends its application, further cementing the PIML model's role as a key tool in hydrological research.